\documentclass{llncs}
\usepackage{graphicx}
\usepackage{enumerate}
\usepackage[cmex10]{amsmath}
\usepackage[colorlinks=true,linkcolor=blue,urlcolor=blue,filecolor=blue,citecolor=green]{hyperref}
\usepackage{amssymb}
\usepackage{amsfonts}
\usepackage{url}
\usepackage{multirow}
\usepackage{fancyhdr}
\usepackage{subcaption}
\usepackage{fullpage}

\newcommand{\stam}[1]{}

\newcommand{\true}{\texttt{true}\:}
\newcommand{\buchi}{B{\"{u}}chi\:}

\newcommand{\globally}{{\bf{G}}\:}

\newcommand{\comment}[1]{}

\newcounter{myctr}

\begin{document}

\title{ARSENAL: Automatic Requirements Specification Extraction from Natural Language}

\author{Shalini Ghosh$^1$, Daniel Elenius$^1$, Wenchao Li$^1$, \\ Patrick Lincoln$^1$, Natarajan Shankar$^1$, Wilfried Steiner$^2$}
\institute{$^1$CSL, SRI International, Menlo Park.\quad \{shalini,elenius,li,lincoln,shankar\}@csl.sri.com\\
$^2$TTTech C. AG, Chip IP Design, Austria.\quad wilfried.steiner@tttech.com
}

\date{}

\maketitle

\begin{abstract}
Requirements are informal and semi-formal descriptions of the expected
behavior of a complex system from the viewpoints of its stakeholders
(customers, users, operators, designers, and engineers).  However, for
the purpose of design, testing, and verification for critical systems,
we can transform requirements into formal models that can be analyzed
automatically.  ARSENAL is a framework and methodology for
systematically transforming natural language (NL) requirements into
analyzable formal models and logic specifications.  These models can
be analyzed for consistency and implementability.  The ARSENAL
methodology is specialized to individual domains, but the approach is
general enough to be adapted to new domains.

\end{abstract}

\section{Introduction}
\label{sec:intro}
Natural language (NL) processing and understanding is becoming
increasingly important in the field of requirements
engineering. Requirements specify important properties of software
systems, e.g., conditions required to achieve an objective, or desired
system invariants.  Requirements in formal languages are precise
and useful for checking consistency and verifying properties, but can
be cumbersome to specify. As a result, stakeholders often prefer
writing NL requirements --- these can be written easily without burden
of formal rigor, but can be inherently imprecise, incomplete, and
ambiguous. NL descriptions and formal modeling languages each offer
distinct advantages to the system designer --- we aim to leverage the
best of both, to aid the system designer in coming up with the
first-cut of a formal model from NL requirements in an automated
fashion. This model can then be refined through iterations with the
human in the loop -- this could enable cost reduction in system
design, while providing high levels of assurance for critical
systems. The main objective of this paper is to answer the question:
\begin{small} 
{\bf {\em ``Can we design such a methodology that
      combines the strengths of natural and formal
      languages for requirements engineering?''}}
\end{small}
With this goal in mind, we present the methodology of ARSENAL:
``\underline{A}utomatic \underline{R}equirements
\underline{S}pecification \underline{E}xtraction from
\underline{Na}tural \underline{L}anguage'.

\begin{figure*}[htbp]
\centering
\fbox{\includegraphics[width=0.7\textwidth]{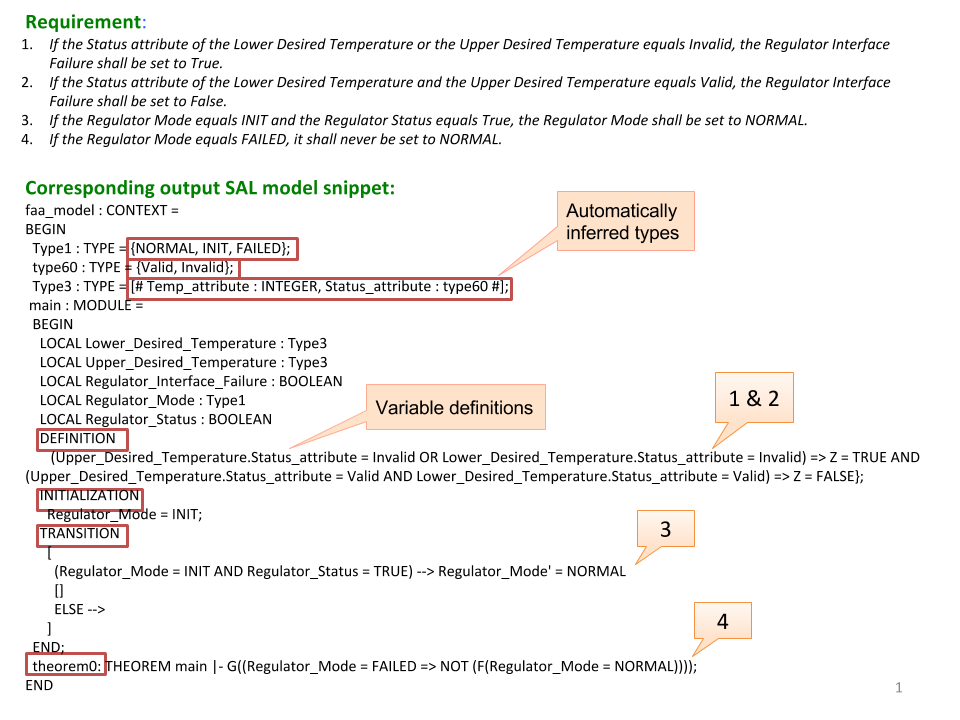}}
\caption{Example showing a SAL Model snippet created from a set of NL
  requirements. Numbers annotating different parts of the SAL model
  correspond to the numbers of the requirements sentences.  }
\label{fig:sal-example}
\end{figure*}

In this paper, we focus on mapping NL requirements to transition
systems expressed in SAL and logic specifications in Linear Temporal
Logic (LTL), for safety critical systems.  SAL~\cite{sal} is a formal
language for specifying transition systems in a compositional way
(details in Section~\ref{sec:sal}), while LTL~\cite{ltl} is a logic
for describing temporal properties.  Figure~\ref{fig:sal-example}
shows a set of requirements from the FAA-Isolette domain and the SAL
model snippet automatically generated by ARSENAL --- it shows how
different parts of the SAL model are constructed from different
requirements sentences. This is a simple model generated automatically
by ARSENAL from 4 requirements sentences --- in some of our domains,
ARSENAL was able to generate a full model from 30+ requirements
sentences in an automated fashion, which is a non-trivial task even
for expert human users.  ARSENAL is able to create a complete formal
model from a set of NL requirements sentences by using a combination
of approaches: preprocessing, type rules, intermediate language
representation, output adapters, and powerful formal methods tools.

ARSENAL has two stages --- Section~\ref{sec:nlp} gives an overview of
the Natural Language Processing (NLP) stage, while
Section~\ref{sec:fm} gives an overview of the Formal Methods (FM)
stage.  We evaluate ARSENAL in Section~\ref{sec:result} and conclude in Section~\ref{sec:future}.

\section{Natural Language Processing}
\label{sec:nlp}
\begin{figure}[h!]
\centering
\includegraphics[width=10cm]{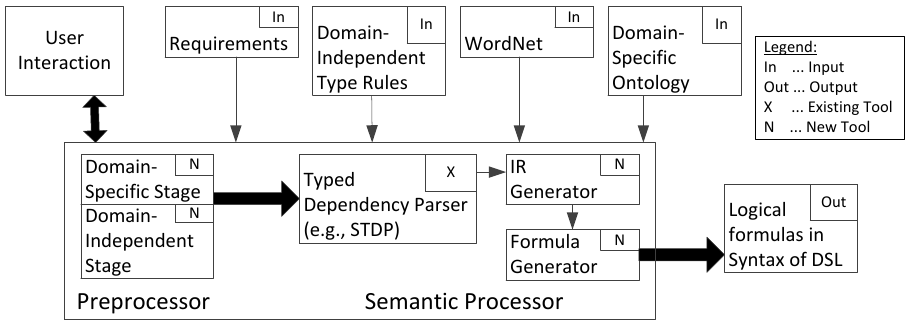}
\caption{NLP Stage of ARSENAL pipeline.}
\label{fig:nlp-stage}
\end{figure}

The NLP stage takes NL requirements as input and
generates a set of logical formulas as output.  The different
components of the NLP stage (Figure~\ref{fig:nlp-stage}) are described
in this section.

\subsection{Preprocessor} 

The first part of the NLP stage is a preprocessor, which enables the
use of a general-purpose semantic parser downstream. It uses both
domain-specific and domain-independent preprocessing. Domain-specific
preprocessing includes identifying entity n-grams like ``Lower Desired
Temperature'' (based on an input ontology) and converting them to
terms like \texttt{Lower\_Desired\_Temperature}. Domain-independent
preprocessing tasks include identifying arithmetic expressions, e.g.,
replacing ``[x + 5]'' by \texttt{ARITH\_x\_PLUS\_5}, so that the
parser treats this as one instead of five terms. The preprocessor also
encodes complex phrases like ``is greater than or equal to'' as
simpler terms like \texttt{dominates}, which can be decoded in the
succeeding FM stage as necessary.

\subsection{Stanford Typed Dependency Parser} 

The next part of the NLP stage is the application of the Stanford
Typed Dependency Parser (STDP)~\cite{stdp} to the preprocessed
sentence --- it generates a set of typed dependency (TD) triples,
which encode the grammatical relationship between mentions (i.e.,
unique entities) extracted from a sentence. TDs are of the form:
\texttt{relation (governor, dependent)}. For example, consider:

\noindent \texttt{prep\_of(Status\_attribute-3, Lower\_Desired\_Temperature-6)}

The above TD indicates that the mention \texttt{Status\_attribute} is
related to
\texttt{Lower\_Desired\_Temperature} via the prepositional connective 
``of''. The suffix of each mention is its position index in the
sentence, which helps to uniquely identify the mention if it has
multiple occurrences in the sentence.  We will denote governor and
dependent terms by \texttt{?g} and
\texttt{?d} respectively.

Let us consider the following example requirements specification
sentence for a temperature regulator in an isolette (an infant
incubator providing controlled temperature, humidity, and oxygen):

\noindent {\bf REQ1:} {\em If the Status attribute of the Lower Desired
  Temperature is invalid, the Regulator Interface Failure shall be set
  to True.}

\begin{figure}
\centering
\begin{minipage}[b]{.5\textwidth}
\centering
\includegraphics[width=\textwidth]{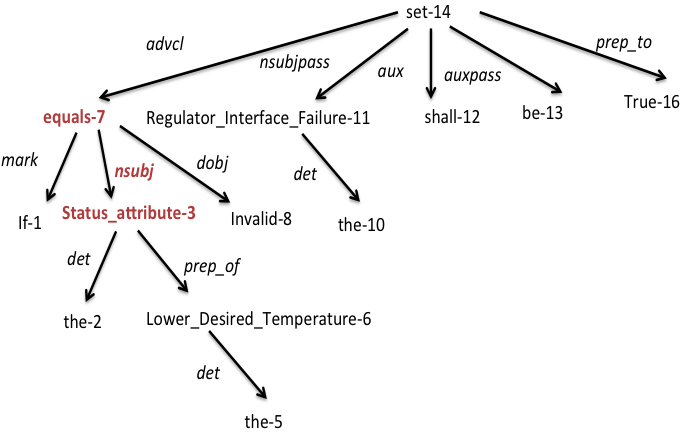}
\caption{Dependencies generated using STDP.}
\label{fig:nlp-stdp}
\end{minipage}
\hfill
\begin{minipage}[b]{.45\textwidth}
\centering
\includegraphics[width=\textwidth]{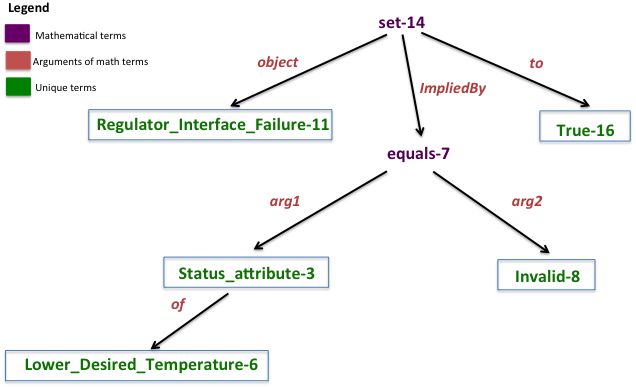}
\caption{Predicate graph after applying the type rules.}
\label{fig:nlp-predicate-graph}
\end{minipage}
\end{figure}

\noindent For REQ1, the full set of TDs generated by STDP maps straightforwardly
to a directed graph rooted at \texttt{set-14}, as shown in
Figure~\ref{fig:nlp-stdp}.

\subsection{Semantic Processor}

\begin{figure}[h!]
\centering
\fbox{\includegraphics[width=0.7\columnwidth]{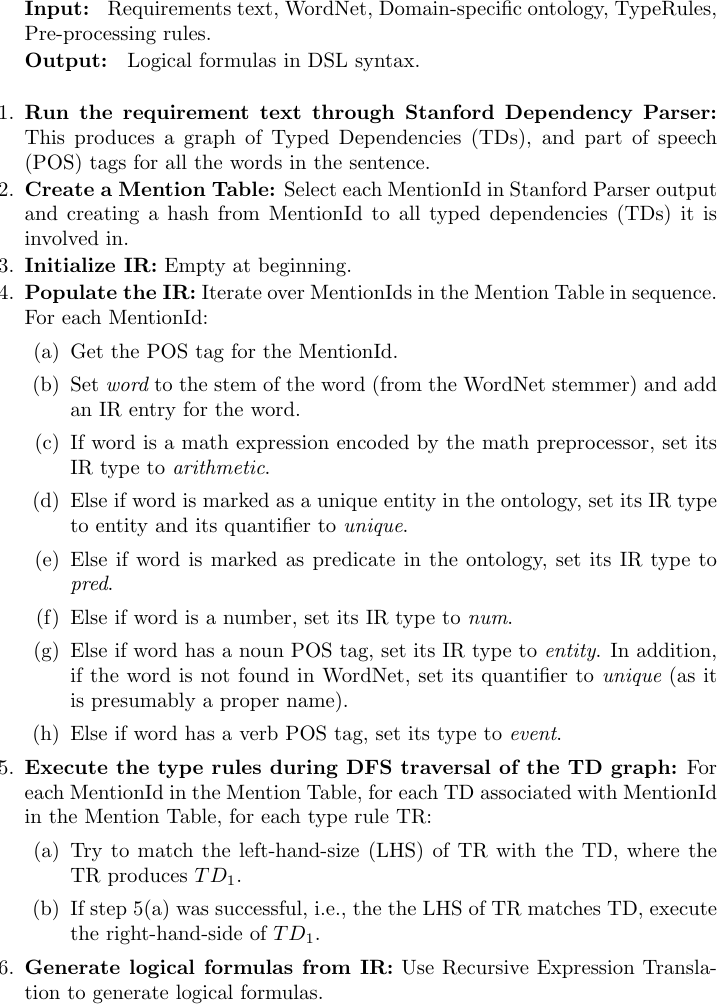}}
\caption{Detailed algorithmic flow of NLP stage.}
\label{fig:nlp-main}
\end{figure}

One key challenge we face in ARSENAL is the mapping from 
NL sentences to LTL formulas in a way that the LTL semantics are
correctly preserved.  The output of STDP is a set of grammatical
relations that lack semantic meaning --- we developed a semantic
processor that takes the output from STDP and systematically applies a
set of type rules to the mentions and dependencies to associate
meanings to them. Each type rule specifies a mapping from a set of
dependencies (grammatical relations between mentions) to a set of
predicates with built-in semantics.  The semantic processor generates
an Intermediate Representation (IR) by annotating the output of STDP
with metadata tags and by the consecutive application of type
rules. The overall algorithm for generating logical formulas from NL
requirements is outlined in Figure~\ref{fig:nlp-main}, while metadata
tags and type rules are described in the rest of this section.

\subsubsection{Metadata tags}
Different types of metadata tags are used to annotate the IR table entries,
indicating if terms are entities/events/predicates, if terms are negated,
quantifier types, temporal relations, etc.

These tags are used to associate semantics with entries in the IR
table. The metadata tags are similar to role annotations in automatic
semantic role labeling~\cite{connor:asrl12}. In this stage, ARSENAL
uses WordNet~\cite{wordnet} and a Part-of Speech (POS) tagger to
identify word stems and to infer unique proper nouns.  ARSENAL can
also use special domain-specific ontologies or glossaries to annotate
the IR entries with a richer set of metadata tags, which can be used
in downstream processing.

\subsubsection{Type Rules} 

The majority of type rules are domain-independent semantic rules used
by the semantic processor to create the IR table --- each type rule
specifies a mapping from a set of dependencies to a set of relational
predicates, with built-in semantics. For example, \texttt{nsubjpass(V,
N)} in the STDP output indicates that the noun phrase \texttt{N} is
the syntactic subject of a passive clause with the root
verb \texttt{V} .  The type rule corresponding to the
TD \texttt{nsubjpass(V, N)} indicates that \texttt{N} is a who/what
argument of relation \texttt{V} in the output formula. Type rules have
the form: \texttt{TD(arg1,arg2): ACTION(arg3,arg4)}. For
example: \texttt{prep\_upon(?g,?d): implies(?d,?g)}\\

\noindent {\bf Matching of Typed Dependencies with Type Rules:} Matching this
rule with the TD \texttt{prep\_upon(entering-17, set-4)} produces a match with
\texttt{?g = entering-17, ?d = set-4}, with an action
\texttt{implies(set-4, entering-17)}. The \texttt{implies(?d,?g)} action adds 
\texttt{impliedBy:?d} to the IR entry for \texttt{?g}.

ARSENAL has various type rules, e.g., for handling implications,
conjunctions/disjunctions, universal/existential quantifiers, temporal
attributes/relations, relation arguments, events.\\

\noindent {\bf Rules with complex patterns:} Some type rules 
have multiple TDs or conditions that match on the left-hand side. For
example: \texttt{nsubj(?g,?d) \& event(?g): rel(agent,?g,?d)}

\begin{figure}[hbtp]
\centering
\fbox{\includegraphics[width=0.7\columnwidth]{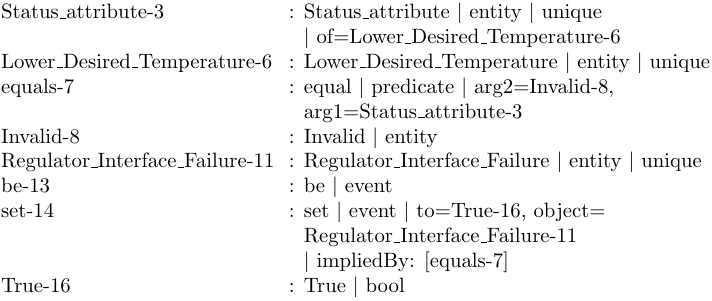}}
\caption{IR table for REQ1.}
\label{fig:ir-example}
\end{figure}

Figure~\ref{fig:ir-example} shows the IR table for REQ1 after
application of the type rules. Currently the type rules are created by
domain experts, aided by a statistics generator that finds the
dominant dependencies in a corpus.  However, most type rules are
domain-independent and can be re-used in other domains --- the
domain-dependent ones need to be customized by tuning based on a
training corpus\footnote{The set of type rules used by ARSENAL are
  available at:
  \url{http://www.csl.sri.com/users/shalini/arsenal/type_rules.txt}.}.
We expect that the set of type rules will evolve towards
quasi-completion with the adoption of ARSENAL in cross-industry
domains.

\subsection{Formula Generation}

There are multiple {\em output adapters} in ARSENAL, which convert the
IR table (with the semantic annotations) to different output forms,
e.g., first-order logic (FOL) and linear temporal logic (LTL)
formulas. In this paper, we discuss the SAL model adapter, which
converts the IR table to LTL formulas that are used in creating a
downstream SAL model. An LTL formula is interpreted over infinite
traces and is built from atomic propositions, Boolean operators
(i.e. negations, conjunctions and disjunctions), and temporal
operators (e.g., $\globally$ operator for always globally true, in
every point of a trace). ARSENAL uses Recursive Expression Translation
(RET) rules to generate the LTL formulas from the predicate graph. For
example, to generate an LTL formula from the predicate graph shown in
Figure~\ref{fig:nlp-predicate-graph}, we can employ the subset of RET
rules in Table~\ref{table:nlp-formula-rules}.

\begin{table*}
\centering
\caption{Partial list of Recursive Expression Translation Rules}
\label{table:nlp-formula-rules}
\begin{tabular}{|c|c|}
\hline
Predicate & Expression Translation\\
\hline \hline
{\em unique}($X$) & $tr^u(e(X)): e(X)$\\
{\em of}($X$, $Y$) & $tr^l(e(X)): tr^m(e(Y)).tr^m(e(X))$\\
{\em set}($X$) $\wedge$ {\em arg1}($X$, $Y$) $\wedge$ {\em arg2}($X$, $Z$) & 
$tr^m(e(X)): tr^u(e(Y)) = tr^u(e(Z))$\\
{\em equal}($X$) $\wedge$ {\em arg1}($X$, $Y$) $\wedge$ {\em arg2}($X$, $Z$) & 
$tr^m(e(X)): tr^u(e(Y)) = tr^u(e(Z))$\\
{\em impliedBy}($X$, $Y$) & $tr^l(e(X)): tr^m(e(Y)) \rightarrow tr^m(e(X))$\\
\hline
\end{tabular}
\end{table*}

We use $e(X)$ to denote the expression associated with mention $X$,
which we will repeatedly rewrite during the translation process.  If
mention $X$ is associated with a unique term, i.e., {\em unique}($X$),
its expression is simply the English word in the mention, e.g.,
$e(\text{Invalid-8}) = \text{Invalid}$.  Given a predicate graph, we
recursively apply the translation rules starting from the root
($\text{set-14}$ in this case). The superscripts $u$, $m$ and $l$
indicate the types of the translation rule for unique terms, simple
arithmetic expressions and logical expressions respectively. When
multiple rules are applicable to the same mention, they are applied in
the order of $tr^l$ followed by $tr^m$ and then $tr^u$. FORM1, the
resulting LTL formula of REQ1 after applying the translation rules, is
shown below:\\

\begin{small}
\noindent {\it {\bf FORM1: G}}\texttt{((Lower\_Desired\_Temperature\\
.Status\_attribute = Invalid) => Regulator\_Interface\_Failure = TRUE)}\\
\end{small}

\noindent The recursive expression translation is equivalent to running a DFS
traversal on the predicate graph with emissions. Every requirement is
considered to be {\em global} except when certain indicative words
such as ``initialize'' are present in the sentence.  
Since $\globally p$ holds of a computation when
condition $p$ holds of all suffixes of the computation,
we have the $\globally$ operator in front of the formulas.

\subsection{Graph Transformations: Dependency Graph $\rightarrow$ Predicate Graph $\rightarrow$ Formula}

The NLP stage can be considered as a sequence of graph
transformations.  The result of applying STDP can be represented as a
dependency graph, which is a directed graph that shows the type
dependencies (Figure~\ref{fig:nlp-stdp}). The IR table can be
considered to be an annotated adjacency list corresponding to the
dependency graph, where additional annotations are stored in the IR
table along with the type dependency information --- these annotations
correspond to metadata generated by type rules.
%and are used in down-stream processing.  
During generation of the logical formula
FORM1, the dependency graph is transformed to a predicate graph. The
predicate graph selects and refines the relations in the IR that are
relevant for the output language. The predicate graph for REQ1 is
shown in Figure~\ref{fig:nlp-predicate-graph}, where the edges
represent binary predicates (with predefined meanings).  The unique
unary predicates are indicated by boxes in the figure.  Additionally,
mentions of indicative words such as ``equals'' and ``set''
are associated with predefined predicates
\textit{equal} and \textit{set}. For example, the predicate \textit{set} 
means that its first argument (object) is a variable being set to a
value which is its second argument (to). The semantic information in
the predicate graph is further interpreted in the FM stage, based on
the target language and additional information about the model.

The NLP stage interprets the NL of the input specifications in the
context of an application domain and usages that are specific to this
domain.  This provides several constraints that give useful context
for narrowing different interpretations of the NL, which enables the
user to generate NL input specifications without restricting the modes
of expression to particular templates.

\section{Formal Analysis}
\label{sec:fm}
In this section, we discuss the Formal Methods
(FM) stage in detail.
%as shown in Figure~\ref{fig:fm-stage}. 
The overall flow of the FM stage is shown in Figure~\ref{fig:fm-stage}
-- this stage takes as input a set of logical formulas as generated by
the NLP stage and creates a composite formal model. The FM stage
uses a combination of consistency, satisfiability and realizability
checks to formally validate the completeness/correctness of the
requirements.

\subsection{Consistency Analysis}

Requirements are as error-prone as implementation, and can themselves
be inconsistent.  In our context, inconsistencies can arise from human
errors in writing the NL requirements or from inaccuracies introduced
by ARSENAL.  Given a set of requirements formalized as LTL formulas,
we check if there exists a {\it model} for the formulas, i.e., they
are {\it satisfiable}.  If the formulas are {\it unsatisfiable}, then
it can be due to errors in the specification or due to errors in the
NLP stage -- in our experiments, we empirically evaluate the
robustness of the NLP stage, to determine how much of the
error can be attributed to the NLP stage in ARSENAL.

\begin{figure}[tbp]
\centering
\includegraphics[width=9cm]{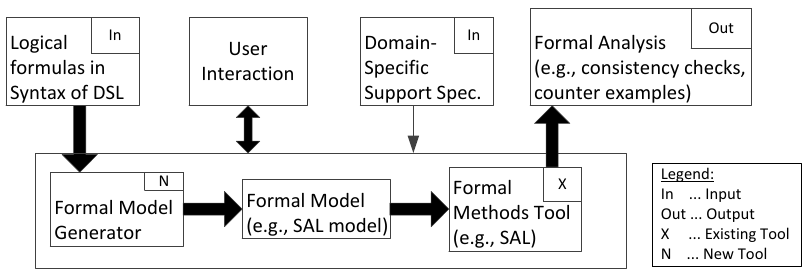}
\caption{FM Stage of ARSENAL pipeline.}
\label{fig:fm-stage}
\end{figure}

The problem of {\it LTL satisfiability checking} can be reduced to
checking emptiness of \buchi automata~\cite{vardi:96}.  Given an LTL
formula $\phi$, we can construct a \buchi automaton $A_\phi$ such that
the language of $A_\phi$ is exactly equivalent to the model of
$\phi$~\cite{vardi-94}.  If the language of $A_\phi$ is empty then
$\phi$ is unsatisfiable, indicating an inconsistency in the
requirements --- we report this inconsistency to the ARSENAL end-user.
Otherwise, we proceed to creating the SAL model, as shown in
Figure~\ref{fig:fm-stage}.

\subsection{Model Checking}
\label{sec:sal}

A SAL model~\cite{sal} represents a transition system whose semantics
is given by a Kripke structure, which is a kind of nondeterministic
automaton widely used in model checking~\cite{modelchecking}. SAL can
be used to prove theorems by encoding properties about the
requirements and using bounded model-checking.  The SAL model-checkers
use LTL --- this is an appropriate language for formally expressing
requirements, since many requirements contain notions of temporal
relations, e.g., eventually (F), always (G).

A transition system like SAL is composed of modules, where each module
consists of a {\it state type}, an {\it invariant definition}, an {\it
initialization condition}, and a {\it binary transition relation} on
the state type. Creating a complete SAL model directly from text with
correct semantics is non-trivial, since formulas need to be
categorized as definition, initialization, transition, or
theorem. Each variable in SAL is also {\it typed}, which needs to be
explicitly specified. During the model generation stage, ARSENAL
gathers type evidences for each variable across all sentences, and
then performs type inference by merging them into equivalence classes
(details of the type merging algorithm are omitted here due to lack of
space). Further, in case of a type conflict, an inconsistency warning
is generated, thus helping the user to refine their NL requirements at
an early stage.

When the model does not satisfy the specification, a negative answer
(often in the form of a counterexample) is presented to the user as a
certificate of how the system fails the specification --- if SAL finds
a counterexample, we know the property encoded in the requirements
does not hold. If SAL does not find a counterexample at a known depth
of model-checking, we next try to see if the LTL formulas are
realizable. Once the specification becomes realizable, an
implementation can be generated automatically, e.g., in Verilog\footnote{An example of a synthesized Verilog model for the FAA domain is available at:
  \url{http://www.csl.sri.com/users/shalini/arsenal/faa-isolette.v}}.

\subsection{Temporal Logic Synthesis}

Given an LTL specification, it may also be possible to directly {\it
synthesize} an implementation that satisfies the specification.  It
has been shown that a subclass of LTL, known as Generalized Reactivity
(1) [GR(1)], is more amenable to synthesis~\cite{gr1-syn} and is also
expressive enough for specifying complex industrial designs.  
We have
incorporated into ARSENAL a counterstrategy-guided assumption mining
approach developed for GR(1) LTL formulas~\cite{li-memocode11}, which
allows adding assumptions to the formulas until either the
specification is realizable or all the recommendations are rejected by
the user.

\section{Evaluation}
\label{sec:result}
In this section, we present results on analyzing ARSENAL's ability in
handling complex NL sentences and different corpora.  We measure 
(a) the robustness of ARSENAL to noise, and
(b) the accuracy of the NLP stage.

\subsection{Degree of Perturbation Metric}
\label{sec:perturb}
\begin{table*}[tbp]
\centering
\caption{Results of perturbation test on ARSENAL.}
\begin{tabular}{|c||c|c|c||c|c|c|}
\hline
Perturbation&\multicolumn{3}{c||}{TTEthernet domain (TTE)}&\multicolumn{3}{c|}{FAA-Isolette domain (FAA)}\\
\cline{2-7}
Type & Total & Perturbed & Accuracy & Total & Perturbed & Accuracy \\
     & sentences & sentences & & sentences & sentences & \\
\hline\hline
And\ $\rightarrow$\ Or       & 36 & 16  & 87\%  & 42 & 13  & 92\%   \\
Is\ $\rightarrow$\ (Is not)    & 36 & 17  & 100\% & 42 & 13  & 92\%   \\
(If A then B)\ $\rightarrow$\ (B if A) & 36 & N/A & N/A   & 42 & 40  & 65\%   \\
\hline
\end{tabular}
\label{tab:perturb}
\end{table*}

We define an evaluation criteria for measuring the robustness of
ARSENAL --- if modifications are made to a requirements sentence using
some rewrite rules, can ARSENAL still generate the right output
formula?  For the FAA and TTE datasets, we perturbed the requirements
using a few transform operators --- the results are outlined in
Table~\ref{tab:perturb}. Let us consider the first row of the table --
the operator modifies all occurrences of ``and'' in a sentence with
``or''. The ``Perturbed sentences'' column indicates how many
sentences in the corpus were affected by applying this operator, and
``Accuracy'' indicates the percentage of perturbed sentences where the
output formula was correct.

\subsection{NLP Stage Accuracy}
\label{sec:levenshtein}
Our goal here was evaluating the accuracy of the NLP stage of
ARSENAL. There are existing metrics to calculate the degree of overlap
between two semantic feature structures, e.g., Smatch~\cite{smatch}
uses ILP and hill-climbing. For ARSENAL, we took the more direct
approach of estimating how many sub-formulas are inserted, deleted or
modified by ARSENAL in the NLP stage, while generating the output
formula for a requirements sentence. We considered a {\em ground-truth
corpus} (requirements sentences annotated with the expected output
formula) and used the following algorithm:

1. Given each requirements sentence in this ground-truth corpus,
generate all sub-formulas $G$ of the ground-truth formula and all
sub-formulas $A$ of the formula generated by ARSENAL.

2. For each sub-formula in $A$, find the best-matching sub-formulas in
  $G$ using a relaxed version of {\it max-weighted matching in
  bipartite graphs} -- the match score between formulas is calculated
  using Typed Levenshtein distance, a variant of the standard
  Levenshtein edit distance, which we designed (details below).

3. Calculate precision, recall and F-measure using similarity
between matched pairs of sub-formulas.

\begin{figure}[hbtp]
\centering
\fbox{\includegraphics[width=0.7\columnwidth]{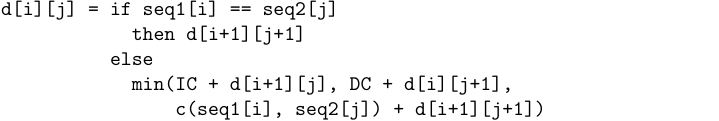}}
\caption{Computation of Typed Levenshtein distance.}
\label{fig:lev}
\end{figure}

Typed Levenshtein Distance is a modification of Levenshtein string
edit distance for formulas, which considers distances between logical
symbols, variables and string tokens differentially. Figure~\ref{fig:lev}
outlines the computation --- here, d is the Typed Levenshtein
distance, IC is insertion cost, DC is the deletion cost, seq1 and seq2
are the two sequences, and c(A, B) is the cost of replacing A by B. We
consider that each formula can have 3 types of tokens:
(1) LogicalSymbol Token (e.g., ``and'', ``or''): c(A,B) = 1 if A $\neq$ B;
(2) String Token (e.g., argument name): c(A,B) = LevenshteinStringEditDistance(A,B); 
(3) Variable Token: c(A,B) = 0, since there should be no cost for changing variable names.

We evaluated the F-measure of the NLP stage of ARSENAL on sections in
the TTE requirements~\cite{steiner:10}. We also ran the evaluation on
requirements from the GPCA, LTLMop and Eurail requirements
documents. We trained ARSENAL (e.g., tuned the preprocessing rules) on
some of the requirements sentences, and evaluated the F-measure only
on the hold-out test set on which ARSENAL was not trained in any
way. Considering the different corpora, there were overall 300
sentences --- of these, we used 112 sentences as the training set for
tuning ARSENAL (from the FAA and TTE requirements), and evaluated the
performance of ARSENAL on the remaining 188 sentences that were held
out.  Table~\ref{tab:fmeasure} shows the results --- on the test set
of 188 requirements, the NLP stage had an overall F-measure of
0.63. Note that none of the requirements from the EUR, GPCA and LTLMop
corpora were used in tuning ARSENAL, but we get good test set accuracy
on these domains, which shows the portability of ARSENAL.

\begin{table}[ht]
\centering
\caption{NLP stage F-Measure on different test-set corpora.}
\begin{tabular}{|c|c|c|}
\hline
Corpus & Test set size & Test set F-measure \\ \hline \hline
TTE & 72 & 0.65\\
EUR & 11 & 0.58\\
GPCA & 70 & 0.62\\
LTLMop & 35 & 0.64\\ \hline \hline
Total & 188 & 0.63\\ \hline
\end{tabular}
\label{tab:fmeasure}
\end{table}

Here are some example errors made by ARSENAL on different domains: \\

\noindent \underline {A) TTEthernet (TTE) requirements}:

\begin{small}
{\bf REQ:} {\em A supported write access to the AMBA 3 AHB-Lite V1.0 slave interface shall be processed and terminated with response OKAY.}
\end{small}

Since there is no TD between ``write'' and ``access'', everything that
follows in the dependency tree is missing from the formula. This can
be fixed by adding a preprocessing entry that treats ``write access''
as an n-gram in this domain.

\begin{small}
{\bf REQ:} {\em If the ct\_mode entry of a CT ID provided at a critical frame output interface is set to anything but OUT, layer 3 shall set the Fo\_State\_Im flag of the Fo\_State field and drop the current transfer on that interface.}
\end{small}

The phrase ``anything but'', another way of stating a negation, it not currently
handled by ARSENAL --- so the output formula misses a sub-formula. \\

\noindent \underline{B) Eurail requirements:}

\begin{small}
{\bf REQ:} {\em A Radio Infill Unit shall never initiate a communication session.}
\end{small}

The negation is missing from the generated ARSENAL formula, because
our type rules interpret ``never'' as a temporal attribute rather than
a normal negation. We can extend type rules to understand ``never'' as
negation in the correct context.

\subsection{Case Study: FAA-Isolette (FAA)}
\label{sec:faa-casestudy}
\begin{figure}
\centering
\begin{minipage}[b]{.6\textwidth}
\centering
\includegraphics[width=0.8\textwidth]{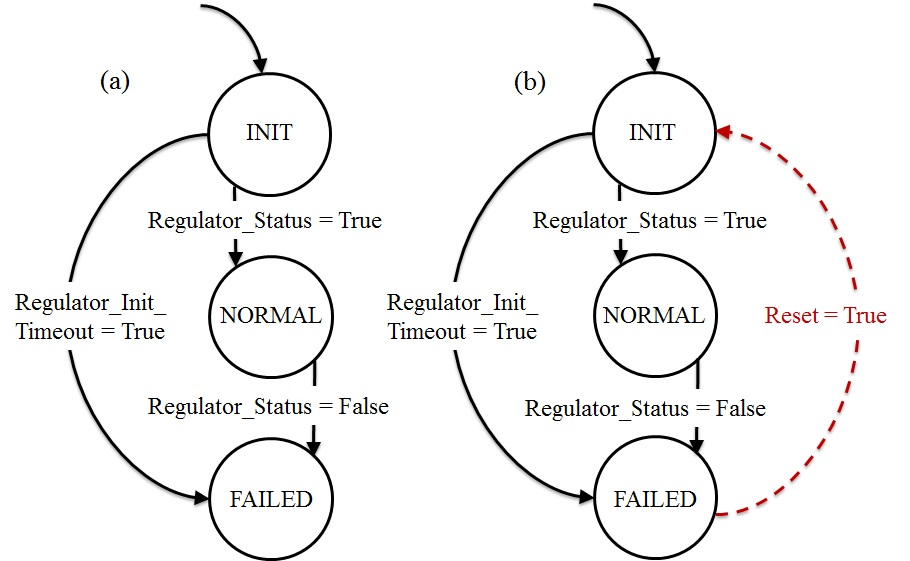}
\caption{Original FSM (a) and Modified FSM (b) for Regulator.}
\label{fig:regulator-fsm}
\end{minipage}
\hfill
\begin{minipage}[b]{.3\textwidth}
 \centering
 \includegraphics[width=0.8\textwidth]{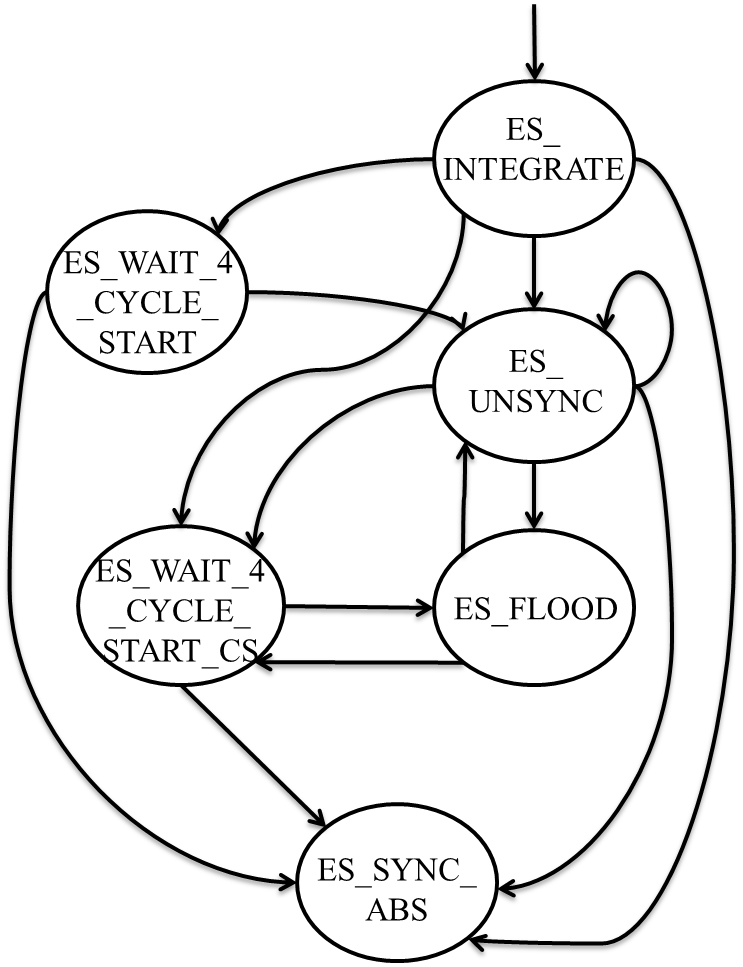}
 \caption{Synchronization FSM in TTEthernet}
 \label{fig:ttethernet}
\end{minipage}
\end{figure}

We also did detailed case studies of the FAA and TTE
requirements\footnote{The requirements corpora for the FAA-Isolette
and TTEthernet domains, and the corresponding SAL models generated by
ARSENAL, are available
at: \url{http://www.csl.sri.com/users/shalini/arsenal/}.} --- we start
with a discussion of FAA requirements. Figure~\ref{fig:regulator-fsm}
(a) shows one of the finite state machines corresponding to the
regulator function of the isolette. An example requirements sentence
is shown below.

\begin{small}
{\bf REQ:} {\it If the Regulator Mode equals INIT and the Regulator Status equals True, the Regulator Mode shall be set to NORMAL.}\\
\end{small}

This experiment seeks to evaluate if ARSENAL can faithfully generate
the transition system corresponding to the description (including the
FSM) in the design document.  Verification was used to validate the
generated FAA model, corresponding to the following SAL theorem
generated by ARSENAL:

\begin{small}
\begin{verbatim}
THEOREM main |- G((Regulator_Mode=FAILED => NOT(F(Regulator_Mode=NORMAL))));
\end{verbatim}
\end{small}

This sentence states that if the FSM is in the FAILED state, then it
cannot go back to the NORMAL state. 
Applying model checking, we verified that the
generated SAL model satisfied the theorem.

We added a sentence corresponding to the transition from FAILED to
INIT state (see Figure~\ref{fig:regulator-fsm} (b)).  For the modified
model, ARSENAL quickly produced a counterexample that showed a path
from the FAILED state to the NORMAL state exists, thus violating the
aforementioned theorem.  This demonstrates that the SAL model
generated automatically by ARSENAL found the injected inconsistency.
After performing satisfiability check, we next tried realizability
check.  Application of LTL [GR(1)] synthesis to these formulas
produced an unrealizable result --- no implementation existed to
satisfy the formulas.  When both \texttt{Regulator\_Status} and
\texttt{Regulator\_Init\_Timeout} are $\true$, the state machine 
(see Figure~\ref{fig:regulator-fsm}) can nondeterministically choose
to go to either the NORMAL state or the FAILED state, from INIT. Such
behavior is not desirable in an actual implementation, so this
specification is not realizable. ARSENAL produced the following
candidate assumption to make the specification realizable.
\begin{small}
\begin{verbatim}
G !(Regulator_Status=1 & Regulator_Init_Timeout=1);
\end{verbatim}
\end{small}
Once the NL sentences describing the transitions are written
differently, such that \texttt{Regulator\_Status}=1 and
\texttt{Regulator\_Init\_Timeout}=1 are mutually exclusive, the
specification becomes realizable and an implementation can be
generated automatically.

\subsection{Case Study: TTEthernet (TTE)}
\label{sec:tte-casestudy}
In the TTEthernet corpus, we consider the NL requirements that
describe the synchronization state machine in TTEthernet.
Figure~\ref{fig:ttethernet} shows the diagram of this state machine
(conditions for transitions are not shown).  The machine starts at the
\texttt{ES\_INTEGRATE} state, and the \texttt{ES\_SYNC\_ABS} state
indicates that the end-system has synchronized with other systems in
the cluster. 
This corpus contains 36 sentences. ARSENAL can handle complex
requirements sentences, generating the correct formula
automatically. An example, describing part of the behavior in the
\texttt{ES\_UNSYNC} state, is shown below.

\noindent
{\bf REQ3:} {\it When an end system is in ES\_UNSYNC state and receives a coldstart frame, it shall (a) transit to ES\_FLOOD state, (b) set local\_timer to es\_cs\_offset, (c) set local\_clock to $0$, (d) set local\_integration\_cycle to $0$, and (e) set local\_membership\_comp to $0$.}

\begin{figure}[hbtp]
\centering
\fbox{\includegraphics[width=0.7\textwidth]{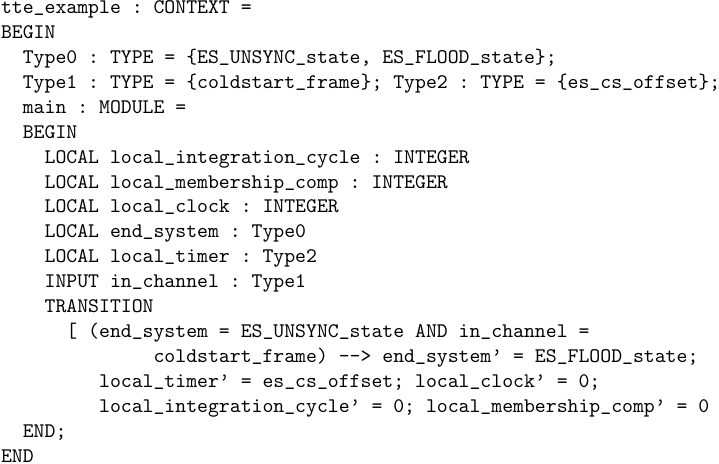}}
\caption{SAL Model for REQ3.}
\label{fig:sal-tte}
\end{figure}

Note that this sentence has a more complicated structure than
REQ1 and includes five itemized actions.  
From the overall SAL model generated automatically, the part
corresponding to REQ3 is shown in Figure~\ref{fig:sal-tte}.  Observe
that ARSENAL was able to infer that the end-system has an enumerated
type (Type0) which contains named values
\texttt{ES\_UNSYNC\_state} and
\texttt{ES\_FLOOD\_state}. 
It was also able to set correctly the type of 
\texttt{local\_integration\_cycle} and \texttt{local\_membership\_comp} to INTEGER.
In this example, the user asserted that all the five LOCAL variables
are {\it state} variables. 
Hence, the actions over these variables were considered as state updates
and mapped to the TRANSITION section. The formula generated by the SAL
adapter corresponding to REQ3 is therefore placed in this section of the SAL model.

A formal method expert was asked to review the model and found it was
compatible with (and in fact, included more information than) a
similar model that he handcrafted in \cite{steiner:10}.  We then
asked one of the original creators of the TTEthernet documentation to
provide a high-level specification that should be verified for this
model.  The sentence in English is given below, followed by the
corresponding LTL theorem in SAL syntax generated by ARSENAL.

\noindent
{\bf REQ4:} {\it If the end system is in ES\_FLOOD state, it shall eventually not be in ES\_FLOOD state.}

\begin{verbatim}
THEOREM main |- G((end_system = ES_FLOOD_state => 
F(NOT(end_system = ES_FLOOD_state))));
\end{verbatim}

We applied bounded model checking, a model checking technique that
checks if the model satisfies the requirement within a bounded number
of transitions, and found a counterexample.  This counterexample
reveals that if the environment keeps sending a
\texttt{coldstart\_frame} to this module, then \texttt{local\_timer},
which maintains a count to timeout in the \texttt{ES\_FLOOD\_state},
will keep resetting to $0$ and thus preventing any transition out of
the \texttt{ES\_FLOOD\_state} to occur.  This helped us identify the
missing assumption (absent in the original documentation) that was
needed for system verification.
In fact, modular verification is one of the most difficult tasks in
verification since it requires the precise specifications of the
constraints on the environment.  These constraints are often implicit
and undocumented.
In this case, the interaction of multiple end-systems should ensure
that any end-system will not receive a \texttt{coldstart\_frame}
infinitely often before it can exit the \texttt{ES\_FLOOD\_state}.

\section{Related Work}
\label{sec:related}
The main advantages of ARSENAL over prior work in requirements
engineering are a less restrictive NL front-end, a more powerful FM
analysis framework, and the ability to generate a full formal model
directly from NL text.

~\cite{kress-gazit:2008}, ~\cite{smith:02} and~\cite{shimizu:02} propose
grammars for representing requirements in template-based natural language.
~\cite{zowghi:01},~\cite{gervasi:05},~\cite{scott:04},~\cite{xiao:sigsoft2012},
~\cite{palsberg:11}, ~\cite{rolland:92}, ~\cite{ryan:93}, ~\cite{dinesh:06},
~\cite{kof:05}, etc. process stylized NL text using
NLP tools and perform different types of checks
(e.g., consistency, access control) on the requirements. Compared to
these methods, ARSENAL uses more state-of-the-art NLP and FM techniques,
and is able to generate a full formal model from input NL requirements using
minimal user supervision.

~\cite{drechsler:2012},~\cite{soeken:2012}, and~\cite{harris:2012} show different
ways of translating high-level NL requirements to tests
in the behavior-driven development (BDD) framework. ARSENAL is more
general --- instead of considering requirements in the test phase, it
considers NL specifications that can be specified in earlier design
phases.

Attempto Controlled English (ACE)~\cite{schwitter:96},
RECORD~\cite{borstler:96} and T-RED~\cite{boman:96} are interactive
requirements processing tools requiring inputs from domain experts,
and are not as automated as ARSENAL.  Related work in semantic parsing
includes PropBank~\cite{propbank}, which annotates text with specific
semantic propositions, and SEMAFOR~\cite{das:hlt10}, which learns
semantic frames per sentence.  ~\cite{liang:acl2011} introduce a new
semantic representation, dependency-based compositional semantics
(DCS), for learning to map questions to answers. ARSENAL's predicate
graph is more general than DCS, as it has both mention-level and
relation-level semantic annotations.~\cite{chen:aaai11} focus on
parsing NL instructions to get a navigational plan for
routing, and learn a semantic parser for the
task.~\cite{walter:robotics13} have a framework for learning
generalized grounding graphs for learning semantic maps from natural
language descriptions. ARSENAL can handle a more general class of
output models, e.g., SAL. \cite{roth:acl2014} propose a scheme for
generating formulas for question/answering from NL requirements, using
semantic parsing --- in ARSENAL, we built an end-to-end system to
generate a complete model with appropriate semantics from multiple
requirements, and did thorough empirical evaluation.

\section{Conclusion and Future Work}
\label{sec:future}
ARSENAL converts NL requirements to formal models in an automated
fashion, which can be further refined through iterations with a human
in the loop. It provides an NL front-end to formal analysis that is
flexible to adapt to usages in different domains. To that end, ARSENAL
can be an important aid for a system designer in designing
high-assurance systems, while reducing cost in the overall design and
manufacturing process.

In the future, we want to test ARSENAL on other domains, generate
other models (e.g., Markov Logic Networks), and go beyond NL text to
handle flow-charts, diagrams and tables. We would also like to explore
learning, e.g., in the NLP stage we currently create the type rules
manually --- we would like to use a learning algorithm like
FOIL~\cite{quinlan:ml90} or Propminer~\cite{akbik:acl13} to learn type
rules. We would also like to explore active learning for incorporating
user feedback.

\bibliography{arsenal-arxiv-april2016}

\newcommand{\etalchar}[1]{$^{#1}$}
\begin{thebibliography}{dMMM06}

\bibitem[AKM13]{akbik:acl13}
Alan Akbik, Oresti Konomi, and Michail Melnikov.
\newblock Propminer: A workflow for interactive information extraction and
  exploration using dependency trees.
\newblock In {\em Proc. of ACL: System Demonstrations}, 2013.

\bibitem[BGL{\etalchar{+}}00]{sal}
S.~Bensalem, V.~Ganesh, Y.~Lakhnech, C.~Mu\ {n}oz, S.~Owre, H.~Rue\ss,
  J.~Rushby, V.~Rusu, H.~Sa\"{i}di, N.~Shankar, E.~Singerman, and A.~Tiwari.
\newblock An overview of {SAL}.
\newblock In {\em LFM}, 2000.

\bibitem[Bor96]{borstler:96}
J.~Borstler.
\newblock User-centered requirements engineering in record - an overview.
\newblock In {\em Proc. of NWPER}, 1996.

\bibitem[BS96]{boman:96}
T.~Boman and K.~Sigerud.
\newblock Requirements elicitation and documentation using {T}-red.
\newblock Master's thesis, Univeversity of Ume\aa, 1996.

\bibitem[CFR12]{connor:asrl12}
M.~Connor, C.~Fisher, and D.~Roth.
\newblock Starting from scratch in semantic role labeling: Early indirect
  supervision.
\newblock 11 2012.

\bibitem[CGP99]{modelchecking}
E.~M. Clarke, Jr., O.~Grumberg, and D.~A. Peled.
\newblock {\em Model checking}.
\newblock MIT Press, 1999.

\bibitem[CK13]{smatch}
Shu Cai and Kevin Knight.
\newblock Smatch: an evaluation metric for semantic feature structures.
\newblock In {\em ACL}, 2013.

\bibitem[CM11]{chen:aaai11}
David~L. Chen and Raymond~J. Mooney.
\newblock Learning to interpret natural language navigation instructions from
  observations.
\newblock 2011.

\bibitem[DDG{\etalchar{+}}12]{drechsler:2012}
R.~Drechsler, M.~Diepenbeck, D.~Gro{\ss}e, U.~K\"{u}hne, H.~M. Le, J.~Seiter,
  M.~Soeken, and R.~Wille.
\newblock Completeness-driven development.
\newblock In {\em Proc. of ICGT}, 2012.

\bibitem[DJLW06]{dinesh:06}
Nikhil Dinesh, Aravind Joshi, Insup Lee, and Bonnie Webber.
\newblock Extracting formal specifications from natural language regulatory
  documents.
\newblock In {\em Proc. of 5th Intl Workshop on Inference in Computational
  Semantics}, 2006.

\bibitem[DJP11]{palsberg:11}
Z.~Ding, M.~Jiang, and J.~Palsberg.
\newblock From textual use cases to service component models.
\newblock In {\em PESOS}, 2011.

\bibitem[dMMM06]{stdp}
M.-C. de~Marneffe, B.~MacCartney, and C.~D. Manning.
\newblock Generating typed dependency parses from phrase structure parses.
\newblock In {\em LREC}, 2006.

\bibitem[DSCS10]{das:hlt10}
Dipanjan Das, Nathan Schneider, Desai Chen, and Noah~A. Smith.
\newblock Probabilistic frame-semantic parsing.
\newblock In {\em Proc. of HLT}, pages 948--956, 2010.

\bibitem[GZ05]{gervasi:05}
V.~Gervasi and D.~Zowghi.
\newblock Reasoning about inconsistencies in natural language requirements.
\newblock {\em ACM TOSEM}, 14(3), 2005.

\bibitem[Har12]{harris:2012}
I.~G. Harris.
\newblock Extracting design information from natural language specifications.
\newblock In {\em DAC}, 2012.

\bibitem[KGFP08]{kress-gazit:2008}
H.~Kress-Gazit, G.~E. Fainekos, and G.~J. Pappas.
\newblock Translating structured {English} to robot controllers.
\newblock {\em Advanced Robotics}, 2008.

\bibitem[Kof05]{kof:05}
Leonid Kof.
\newblock Natural language processing: Mature enough for requirements documents
  analysis.
\newblock In {\em NLDB}, 2005.

\bibitem[KP03]{propbank}
Paul Kingsbury and Martha Palmer.
\newblock Propbank: The next level of treebank.
\newblock In {\em Proc. of Treebanks and Lexical Theories}, 2003.

\bibitem[LDS11]{li-memocode11}
W.~Li, L.~Dworkin, and S.A. Seshia.
\newblock Mining assumptions for synthesis.
\newblock In {\em MEMOCODE}, 2011.

\bibitem[LJK11]{liang:acl2011}
Percy Liang, Michael~I. Jordan, and Dan Klein.
\newblock Learning dependency-based compositional semantics.
\newblock In {\em Proc. of ACL-HLT}, 2011.

\bibitem[Mil95]{wordnet}
G.~A. Miller.
\newblock Wordnet: A lexical database for {English}.
\newblock {\em Comm. of ACM}, 38, 1995.

\bibitem[MP92]{ltl}
Z.~Manna and A.~Pnueli.
\newblock {\em The temporal logic of reactive and concurrent systems}.
\newblock 1992.

\bibitem[PP06]{gr1-syn}
N.~Piterman and A.~Pnueli.
\newblock Synthesis of reactive(1) designs.
\newblock In {\em Proc. of VMCAI}, 2006.

\bibitem[Qui90]{quinlan:ml90}
J.~R. Quinlan.
\newblock Learning logical definitions from relations.
\newblock {\em Machine Learning}, 5, 1990.

\bibitem[RDKS14]{roth:acl2014}
Michael Roth, Themistoklis Diamantopoulos, Ewan Klein, and Andreas Symeonidis.
\newblock {\em Proc. of ACL Workshop on Semantic Parsing}, chapter Software
  Requirements: A new Domain for Semantic Parsers.
\newblock 2014.

\bibitem[RP92]{rolland:92}
Colette Rolland and C.~Proix.
\newblock A natural language approach for requirements engineering.
\newblock In {\em CAiSE}, 1992.

\bibitem[Rya93]{ryan:93}
Kevin Ryan.
\newblock The role of natural language in requirements engineering.
\newblock In {\em RE}, 1993.

\bibitem[SACO02]{smith:02}
R.~L. Smith, G.~S. Avrunin, L.~A. Clarke, and L.~J. Osterweil.
\newblock {PROPEL}: An approach supporting property elucidation.
\newblock In {\em Proc. of ICSE}, 2002.

\bibitem[SCK04]{scott:04}
W.~Scott, S.~Cook, and J.~Kasser.
\newblock Development and application of context-free grammar for requirements.
\newblock In {\em Proc. of SETE}, 2004.

\bibitem[SD10]{steiner:10}
W.~Steiner and B.~Dutertre.
\newblock {SMT}-based formal verification of a {TTEthernet} synchronization
  function.
\newblock In {\em FMICS}, 2010.

\bibitem[SF96]{schwitter:96}
R.~Schwitter and N.~E. Fuchs.
\newblock Attempto controlled {E}nglish ({ACE}) a seemingly informal bridgehead
  in formal territory.
\newblock In {\em Proc. of JICSLP}, 1996.

\bibitem[Shi02]{shimizu:02}
K.~Shimizu.
\newblock {\em Writing, Verifying, and Exploiting Formal Specifications for
  Hardware Designs}.
\newblock PhD thesis, EE Dept., Stanford Univ., 2002.

\bibitem[SWD12]{soeken:2012}
M.~Soeken, R.~Wille, and R.~Drechsler.
\newblock Assisted behavior driven development using natural language
  processing.
\newblock In {\em Objects, Models, Components, Patterns}, volume 7304 of {\em
  LNCS}. 2012.

\bibitem[Var96]{vardi:96}
Moshe~Y. Vardi.
\newblock An automata-theoretic approach to linear temporal logic.
\newblock In {\em Logics for Concurrency}, volume 1043 of {\em LNCS}. 1996.

\bibitem[VW94]{vardi-94}
Moshe~Y. Vardi and Pierre Wolper.
\newblock Reasoning about infinite computations.
\newblock {\em Inf. Comput.}, 115(1):1--37, November 1994.

\bibitem[WHH{\etalchar{+}}13]{walter:robotics13}
Matthew~R. Walter, Sachithra Hemachandra, Bianca Homberg, Stefanie Tellex, and
  Seth~J. Teller.
\newblock Learning semantic maps from natural language descriptions.
\newblock In {\em Robotics: Science and Systems}, 2013.

\bibitem[XPTX12]{xiao:sigsoft2012}
X.~Xiao, A.~M. Paradkar, S.~Thummalapenta, and T.~Xie.
\newblock Automated extraction of security policies from natural-language
  software documents.
\newblock In {\em SIGSOFT FSE}, 2012.

\bibitem[ZGM01]{zowghi:01}
D.~Zowghi, V.~Gervasi, and A.~McRae.
\newblock Using default reasoning to discover inconsistencies in natural
  language requirements.
\newblock In {\em Proc. of APSEC}, 2001.

\end{thebibliography}
\bibliographystyle{alpha}

\end{document}